\documentclass[journal]{IEEEtran}

\usepackage[utf8]{inputenc}
\usepackage[T1]{fontenc}
\usepackage{siunitx}
\usepackage[colorlinks=true, allcolors=blue]{hyperref}
\usepackage{amsmath}
\usepackage{amsfonts}
\usepackage{soul} 
\usepackage{caption}
\usepackage{subfigure}
\usepackage{booktabs} 

\usepackage{tikz,xcolor}
\definecolor{lime}{HTML}{A6CE39}
\DeclareRobustCommand{\orcidicon}
{
    \begin{tikzpicture}
    \draw[lime, fill=lime] (0,0) circle [radius=0.16] 
    node[white] {{\fontfamily{qag}\selectfont \tiny ID}};    \draw[white, fill=white] (-0.0625,0.095) circle [radius=0.007];    
    \end{tikzpicture}
    \hspace{0mm}}
\foreach \x in {A, ..., Z}{%
    \expandafter\xdef\csname orcid\x\endcsname{\noexpand\href{https://orcid.org/\csname orcidauthor\x\endcsname}{\noexpand\orcidicon}}
}

\begin{document}

\title{Vehicle-road Cooperative Simulation and 3D Visualization System}

\author{Di Wu
}

\markboth{Journal of \LaTeX\ Class Files,~Vol.
}%
{Shell \MakeLowercase{\textit{et al.}}: Bare Demo of IEEEtran.cls for IEEE Journals}

\maketitle
\begin{abstract}
The safety of single-vehicle autonomous driving technology is limited due to the limits of perception capability of on-board sensors.
In contrast, vehicle-road collaboration technology can overcome those limits and improve the traffic safety and efficiency, 
by expanding the sensing range, improving the perception accuracy, and reducing the response time.  
However, such a technology is still under development;
it requires rigorous testing and verification methods to ensure the reliability and trustworthiness of the technology. 
In this thesis, we focus on three major tasks:
(1) analyze the functional characteristics related to the scenarios of vehicle-road cooperations, highlightening the differences between vehicle-road cooperative systems and traditional single-vehicle autonomous driving systems; 
(2) refine and classifiy the functional characteristics of vehicle-road cooperative systems; 
(3) design and develop a simulation system, and provide a visual interface to facilitate development and analysis.
The efficiency and effectiveness the proposed method are verfied by experiments.
\end{abstract}


%
\IEEEpeerreviewmaketitle

\section{Introduction}


The use of autonomous driving to solve traffic congestion and safety accidents is an industry trend. Recent autonomous driving research is mainly focused on improving the intelligence of single-vehicle. However, the single-vehicle autonomous driving system only uses the sensors on its car, which results that there still have blind field outside the sensing area of the system. The blind field leads to dangerous situations such as sensors malfunction, undetectable angles, ghost probes, etc. And these problems are hard to be resolved with the development of single-vehicle autonomous systems. 
The vehicle-road cooperation is considered a new research field in autonomous driving. It uses sensors both on the vehicle and roadside cooperatively. Theoretically, this method can expand the perception range of vehicles, improve the perception accuracy, allow the system to predict the trajectory of surrounding objects in more advance, and improve the system's responsiveness. Besides helping the vehicles get more information, it can also improve the traffic efficiency and reduces pollution, reducing the incidence of accidents \cite{usdot2018}. This significantly improves the safety of autonomous driving \cite{9495943}.

However, the vehicle-road cooperate technology is still in the infancy stage. To test and verify the superiority of the vehicle-road cooperative system, we need to design corresponding test scenarios and cases. One of the major problems is that the development of the vehicle-rood cooperative system is costly. And there still has potential safety problems with conducting field testing as they usually require enough amount of autonomous driving vehicles and extra-large testing fields \cite{hyldmar_fleet_2019}. Field tests reflect the safety capability of the autonomous driving system by setting up the vehicle and the corresponding scenario in a realistic field. There is no doubt that field testing is closer to reality and more reflective of what might happen when an autonomous driving system is operating on real roads. Another approach to facilitating experimental research with less cost is simulation tests  \cite{lan2019evolutionary,lan2019simulated}. Simulating the real sense through software will only need to occupy a room-sized space to place the computer server. A variety of hardware devices can be defined through the software to simulate its digital signal. Specific test scenarios for vehicle-road cooperation can be easily built and kept at a low cost. Therefore, it is a better solution for researchers to build a simulation environment in a virtual environment for vehicle-road cooperative simulation testing.

However, existing simulation software developed for autonomous driving only provide limited support for vehicle-road cooperative capabilities. In our investigation, we don't find any existing open-source simulation tool built for the vehicle-road cooperative system featuring both roadside infrastructures and vehicles with a full vehicle-road cooperative autonomous driving system pipeline. Besides, there's no open-source visualization tool for the researchers to intuitively display and analysis the various data from the vehicle-road cooperative autonomous driving system. The researchers are facing difficulties to find an easy and flexible way to deploy, test, and validate the benefit of vehicle-road cooperative autonomous driving systems in simulation. 

In this paper, we present a Vehicle-road Cooperative Simulation and Visualization System (VCSVS), which aims to meet the requirements for vehicle-road cooperative simulation. It is a generalized open-source system integrated with simulation and visualization tools for vehicle-road cooperate research. It provides a data pipeline through multiple layers to simulate a vehicle-road cooperative autonomous driving system. We provide scenario generation tools to construct various test scenarios conveniently for the researchers. More importantly, the system is designed with separate front and back ends. The default modules in each layer can be easily replaced with the researchers' designs and test them in the scenarios. We select CARLA \cite{dosovitskiy_carla_2017} as our base simulator, which uses Unreal Engine 4 \cite{UE4} to render the world and simulate physics, and simulate the vehicles with road-side infrastructures dynamically. We select AVS \cite{AVS} as our base visualization framework to provide 3D visualization of the vehicle-road cooperative system data.

The rest of this paper is organized as follows. Section II describes the related work. Section III shows the system design and describes the details of each layer. Section IV provides the experiment setup and the discussions of the results. Section V concludes the paper and outlines future works.

\section{Related Work}
\label{sec:related_work}


\subsection{Existing Driving Simulators}
\begin{table}[!hb]
\centering 
\caption{Comparison between VCSVS and existing driving simulators}
\label{simulators}
\begin{tabular}{c|m{4.3em}<{\centering}|m{4em}<{\centering}|m{5.5em}<{\centering}}
\toprule
\textbf{Simulator} & \textbf{Realistic Visuals \& Sensors} & \textbf{Predefined Roadside Devices} & \textbf{Vehicle-road Cooperation} \\
\cline{1-4}
TORCS \cite{wymann_torcs_nodate}        & \checkmark & $\times$ & $\times$ \\
\hline
CoInCar \cite{Naumann2018CoInCarSimAO}  & \checkmark & $\times$ & $\times$ \\
\hline
FLUIDS \cite{zhao_fluids_2018}       & \checkmark & $\times$ & $\times$ \\
\hline
Carla \cite{dosovitskiy_carla_2017}        & \checkmark & $\times$ & $\times$ \\
\hline
Sim4CV \cite{muller_sim4cv_2018}       & \checkmark & $\times$ & $\times$ \\
\hline
Apollo \cite{ap}       & \checkmark & $\times$ & $\times$ \\
\hline
GTA \cite{richter_playing_2017}          & \checkmark & $\times$ & $\times$ \\
\hline
VCSVS (ours) & \checkmark & \checkmark & \checkmark \\
\cline{1-4}
\end{tabular}
\end{table}

Driving simulators have extensively taken part in the development of autonomous driving systems. Recent simulators have provided realistic visuals and sensors for autonomous driving system testing  \cite{lan2016developmentVR,lan2016developmentUAV}, but do not provide tools for simulating and visualizing vehicle-road cooperative systems. Simulators like TORCS \cite{wymann_torcs_nodate}, CoInCar-Sim \cite{Naumann2018CoInCarSimAO}, and FLUIDS \cite{zhao_fluids_2018} focus on interactions between multiple vehicles. These simulators are designed for the study of complex interactions between agents-vehicles. Carla  \cite{dosovitskiy_carla_2017}, Sim4CV \cite{muller_sim4cv_2018}, Apollo \cite{ap}, and GTA \cite{richter_playing_2017} explicitly feature physics modeling and realistic rendering for autonomous driving system simulation. They also provide a set of realistic sensors such as cameras, LiDAR. However, these simulators don't provide a predefined way of roadside devices. They don't consider vehicle-road cooperative systems \cite{lan2018directed}. What we aim is to extend their ability to vehicle-road cooperative simulation.

\subsection{Visualization}

In traditional autonomous research, the perceptual information of different sensors is displayed in different ways \cite{gao2021neat}. OpenCV \cite{opencv_library} can display images covered with detection boxes. Matplotlib \cite{hunter2007matplotlib} is a library of visualization tools in Python. It can create static, animated, and interactive visualizations of data. In OpenPCDet \cite{openpcdet2020}, there is a basic visualization tool that can display LiDAR point clouds and the detection 3D boxes with fixed viewing angle and center. In 3DABMOT \cite{Weng2020_AB3DMOT}, it provides a visualization tool that can cover the origin image with tracking results and package all the frames into a GIF image. These tools are suitable for the research and development of specific algorithms, but for a highly integrated and complex autonomous driving system, we need an integrated visualization to study the impact of each module on the final autonomous driving behaviors. To display the information completely, we need a 3D visualization system with the ability of information files.

\begin{table}[!ht]
\centering \small
\caption{Comparison of different visualization frameworks}
\label{simulators}
\begin{tabular}{m{7em}<{\centering}|m{2.2em}<{\centering}|m{2.3em}<{\centering}|m{4em}<{\centering}|m{4.4em}<{\centering}}
\toprule
\textbf{Visualization Framework} & \textbf{Point Cloud} & \textbf{Image} & \textbf{3D Visualize} & \textbf{Interactive} \\
\cline{1-5}
OpenCV \cite{opencv_library}      & \checkmark & \checkmark & $\times$ & $\times$ \\
\hline
Matplotlib \cite{hunter2007matplotlib}  & \checkmark & \checkmark & $\times$ & $\times$ \\
\hline
OpenPCDet \cite{openpcdet2020}   & \checkmark & $\times$ & $\times$ & $\times$ \\
\hline
3DABMOT \cite{Weng2020_AB3DMOT}     & $\times$ & \checkmark & $\times$ & $\times$ \\
\hline
RVIZ \cite{ros}        & \checkmark & \checkmark & \checkmark & \checkmark \\
\hline
WebVIZ \cite{Webviz}      & \checkmark & \checkmark & \checkmark & \checkmark \\
\hline
AVS \cite{AVS}         & \checkmark & \checkmark & \checkmark & \checkmark \\
\cline{1-5}
\end{tabular}
\end{table}

The combination of ROS and RVIZ \cite{ros} is commonly used to display data in robotics research. ROS is Robot Operating System, which is a software and tool library that can help build robot applications \cite{lan2021learning1,lan2021learning2}. RVIZ is a commonly used visualization application in ROS, it can display rich content, and also can carry out some simple interactions. However, the interface is relatively simple, many options need to be customized, and the fusion display is more difficult. WebViz\ref{webwiz} developed by Cruise \cite{Webviz} can visualize thousands of complex decisions of vehicles. It is based on ROS. Drag and drop ROS package files into WebVIZ, and it can provide immediate visual insight into autonomous driving data. But it only supports ROS format data. Autonomous Visualization System (AVS) developed by Uber \cite{AVS} is a fast, powerful, Web-based 3D visualization framework, which uses a separate front and back end design, with XVIZ as the back end and Streetscape.gl as the front end, which can be used to display autonomous driving data. These frameworks provide better visualization of autonomous driving data. But we still need to extend their function to provide a visualization interface for the vehicle-road cooperative autonomous driving system.

\begin{figure}[htbp]
    \centering
    \subfigure[RVIZ interface]{
        \includegraphics[width=.38\linewidth]{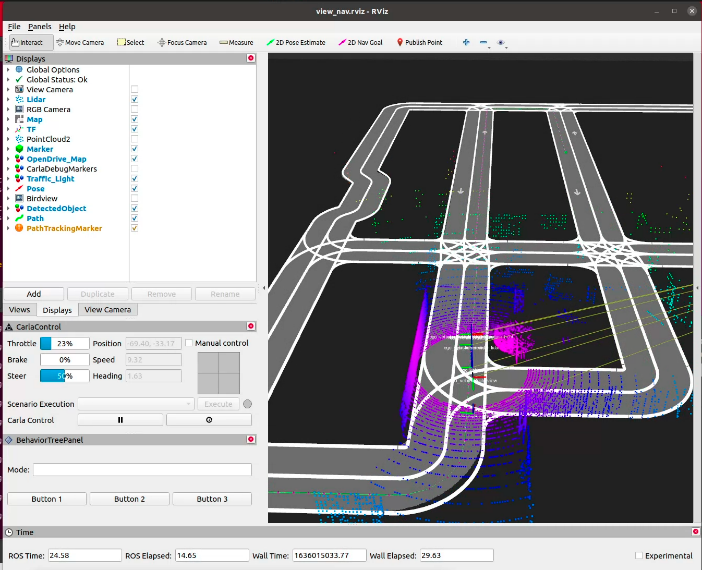}
        \label{rviz}
    }
    \subfigure[WebViz framework]{
        \includegraphics[width=.54\linewidth]{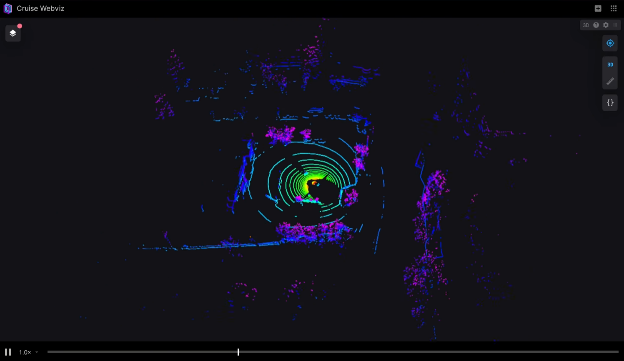}
        \label{webwiz}
    }
    \subfigure[AVS framework]{
        \includegraphics[width=.95\linewidth]{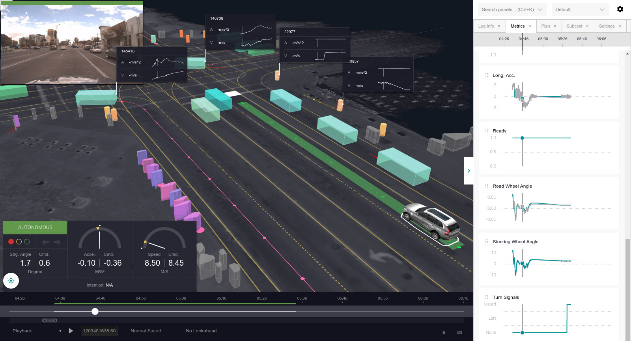}
        \label{avs}
    }
    \caption{Visualization frameworks}\label{}
\end{figure}

\section{Methodology}
\label{sec:methodology}

\begin{figure}[!ht]
    \centering
    \includegraphics[width=0.95\linewidth]{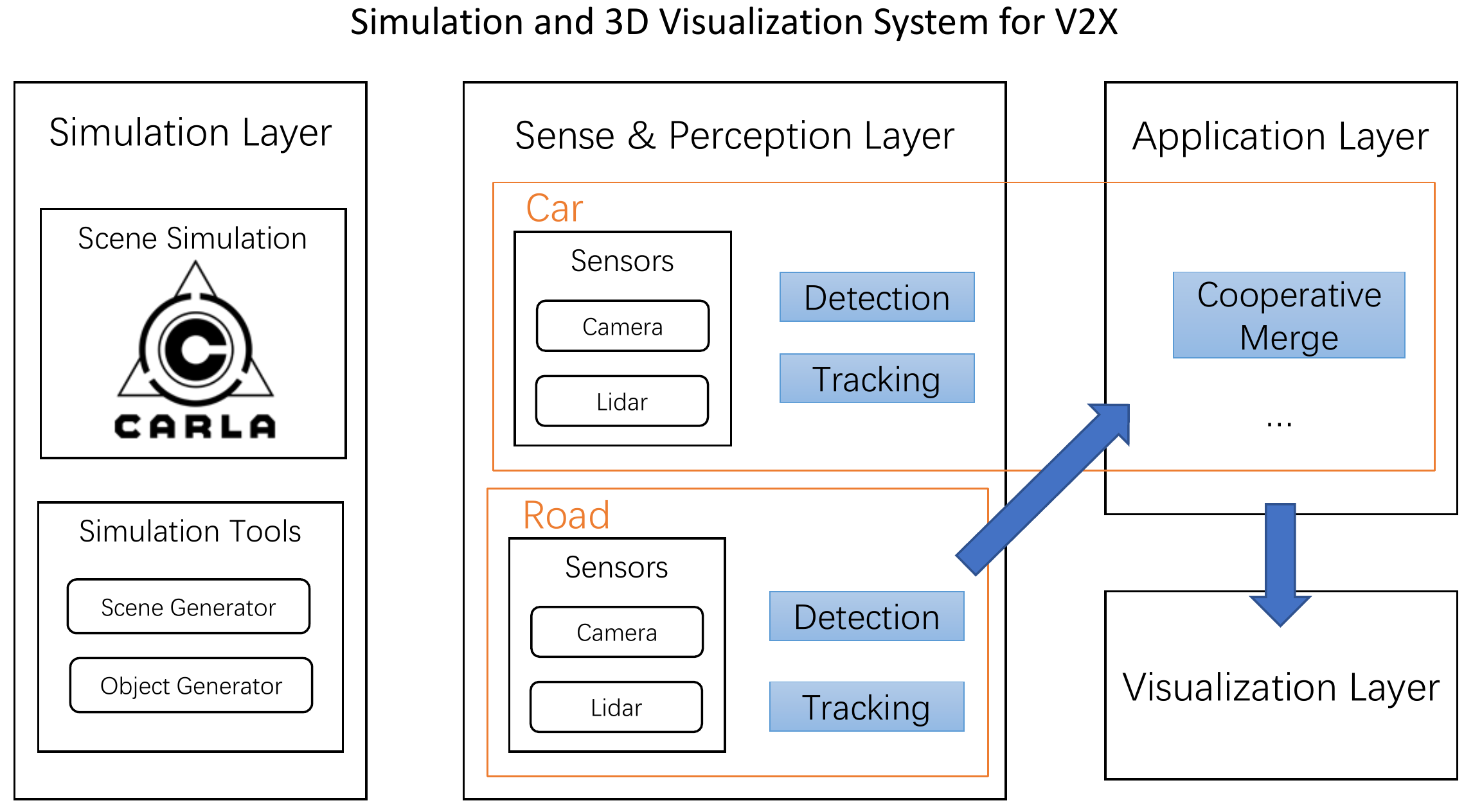}
    \caption{System design of Vehicle-road Collaborative Simulation and Visualization System}
    \label{system}
\end{figure}

The Vehicle-road Cooperative Simulation and Visualization System is a generalized framework integrated with simulation and 3D visualization for dynamic vehicle-road cooperative autonomous driving. It provides simulation tools to generate various vehicle-road cooperative scenarios and a visualization tool for the researchers to deep into the vehicle-road cooperative data. As figure.\ref{system} shows, that the system is composed of four major layers - Simulation Layer, Sense Layer, Application Layer, and Visualization Layer. We specify the data format for each layer as KITTI \cite{Geiger2012CVPR,Menze2015ObjectSF} to ensure data compatibility and data pipeline flow.

\subsection{Simulation Layer}
We use Carla \cite{dosovitskiy_carla_2017} as the basic simulation engine of our system. Carla is a free, open-source autonomous driving simulator designed to promote the development, training, and validation of autonomous driving systems. It uses the Unreal engine \cite{UE4} to achieve high-quality scene rendering, realistic physical effects, and basic sensor modeling. Carla defines a common simulation API that researchers can control all elements in the simulation, from scenario settings to blueprint, as well as sensor placement, testing perception, planning, and control algorithms. The server side will constantly update the physical information of the test scenarios, and the client-side will be controlled by the user through the Carla API. Our simulation system encapsulates the Carla API and provides a higher level of vehicle-road cooperative simulation control. We provide tools for test scenario generation and some basic simulation scenarios for vehicle-road cooperation. We provide various configurations of the scenarios. The researchers can choose a straight road, a curve road, or across as the base scene. Other than the default sunny weather, the researchers can set different weather situations like cloudy, foggy, or rainy. Different vehicles, roadside devices, pedestrians, and bikers can be added to the scenarios as objects. And the researchers can set the initial positions of these objects, as well as their poses and future movement. The simulation world will be controlled by the simulation controller, which will ensure the synchronization of the objects.
\subsection{Sense and Perception Layer}
This layer contains two parts, the extracted sensor information from the Simulation Layer and the perception information of the algorithms on the sensor information. As a vehicle-road cooperative system, each module in this layer will separate into vehicle side and roadside, enabling the researcher to test and compare the difference between vehicle-road cooperative with single-vehicle autonomous driving systems.
\subsubsection{Sensors}
We specifically simulate two kinds of practical equipment for the sensor module, camera, and LiDAR. While the researchers have set the positions and poses of LiDAR and cameras in simulation, they can also set the detection angle, density, and detection range of the LiDAR according to their needs \cite{xu2019online}. For the cameras, they can set resolutions, sensor sensitivity, the field of view, etc. All the sensors can be configured separately. The LiDAR will output the point cloud of each frame in KITTI format. The camera will output frame pictures. We can transfer this data in real-time to other programs or save it to  the disk.
\subsubsection{Detection}
Based on the original data output from the sensors, we can use a variety of object detection algorithms to get the detection information \cite{lan2018real,lan2019evolving}. We mainly focus on vehicles and pedestrian detection as they are important in traffic safety. We choose to identify the targets from the point cloud because we can extract useful 3D object information like positions and poses. We take the OpenPCDet algorithm as default. OpenPCDet \cite{openpcdet2020} is a clear, simple, self-contained open source project for 3D target detection based on LiDAR. Among them, neural network models such as PointRCNN \cite{Shi2019PointRCNN3O} and Voxel R-CNN\cite{deng2020voxel} have good results in 3D target detection and are often used as benchmark methods \cite{lan2017development}. We use the KITTI dataset to test the pre-trained PointRCNN and Voxel R-CNN models and both get good results. The pre-trained PointRCNN model can recognize more types of objects. Therefore, according to our actual settings, we consider using the PointRCNN model as our default for detection.

\subsubsection{Tracking}
After the previous step of detection, we can find targets shown in each frame. And some targets will appear continuously in multiple frames. We can classify these targets and apply the same ID to the same target detected among different frames to facilitate the implementation of target trajectory prediction. For the 3D bounding box obtained by processing point cloud data, we can judge whether it belongs to the same target by its pose relationship between adjacent frames, to track its pose changes in all frames. In the current target tracking methods, the commonly used method is mathematical filtering. We select AB3DMOT \cite{Weng2020_AB3DMOT} as our default algorithm. AB3DMOT is a simple and accurate real-time baseline 3DMOT system, which combines a 3D Kalman filter and Hungarian algorithm for state estimation and data association of 3D bounding boxes. We tested it and made its output into KITTI format.

\subsection{Application Layer}
\subsubsection{Cooperative Merge}
In the vehicle-road collaborative system, the problem we need to consider is how to realize the data sharing between the vehicle and the roadside devices, so that the vehicle can expand its perception range and improve its perception accuracy after getting more information to make a better decision. For the data from the vehicle and roadside devices, there are three common ways of fusion.
\begin{itemize}
    \item Pixel-level fusion
    Pixel-level fusion directly combines the original data collected by the vehicle and the roadside devices. The information will not be lost and the accuracy can be high, but a large amount of data needs to be transmitted. However, in our test, we find the result is not good as we thought. The reason may be that the algorithms are designed only for vehicle side perception. The points outside the vehicle's point cloud may be considered noises.
    \item Decision-level fusion 
    With decision-level fusion, the vehicle and the roadside devices will process and identify the collected data respectively, and only combine the final perception results, which can reduce the communication cost and improve the response speed of the system. But there may be some lost information as there are thresholds when processing the data.
    \item Feature-level fusion
    In the current common neural networks, there usually has a feature layer to extract the feature information from the original data. The size of the features will be smaller than the original data. So we can simply merge the features from the roadside devices to the vehicle to transfer less data while keeping more information. But the neural network model needs to be the same so that the features can be merged.
\end{itemize}

In our system, we set decision-level fusion as default to facilitate processing. From the Sense and Perception Layer, we get 3D boxes of the targets from both the vehicle and roadside devices. The obtained 3D boxes and their labels will be sent to the vehicle. The vehicle autonomous driving system can combine the target information obtained from its perception with the roadside for prediction and decision making. A problem shown here is that the coordinates of the data are relative to the sensor itself. To combine the information, the coordinate system of the data needs to be unified. We set a global virtual coordinate system. For the roadside, we can know the position of the roadside sensor relative to the virtual coordinate system when it is installed, and for the vehicle side, we can calculate the position of the vehicle relative to the global virtual coordinate system at each moment by the GPS location of the vehicle. Then, all the information will be unified in the global virtual coordinate system. And the cooperative merge of information is finished.

\begin{figure}[htb]
    \centering
    \includegraphics[width=.7\linewidth]{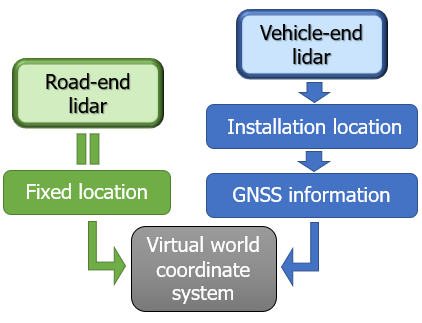}
    \caption{Merge of the coordinate system}\label{}
\end{figure}

\subsection{Visualization Layer}
In addition to providing a basic terminal display, 3D visualization is an important part of our Vehicle-road Collaborative Simulation and Visualization system. It's hard to analyze the information of the autonomous driving system with only the plain text output. The amount of data in each frame may be quite large and there are many algorithm models in the autonomous driving system. We want to have a better visualization of the data.

We implement a 3D visualization interface based on Uber's open-source AVS framework. We visualize the information of each layer from the Simulation System in 3D through filtering, grouping, coordinate fusion, time frame synchronization, and other processing. We load the data and package them into XVIZ format. For each layer, we first parse various formats of raw data into memory. Then we process the data by removing missing items, adding missing values, or filtering noise data. Then we need to synchronize frames of data from each layer by setting the timestamp of each frame. As there are many layers of data, we push them into different XVIZ streams. And for data from the vehicle and roadside devices, we distinguish them by setting different colors. After all, we package them into XVIZ format data.

For the sensors, the laser point cloud will be displayed in the form of a three-dimensional point set. We can use different colors to display the laser point cloud data collected by different ends, vehicles, or roadside devices. The image streams are displayed in separate video windows. They can display the image stream from multiple cameras at the same time. 

For the detection, the detection results for the vehicles and pedestrians will be displayed on the interface in the form of 3D Boxes, and the recognition results of different devices will be displayed in different colors, based on which the researchers can analyze the effect of the object detection algorithm, the detection results of the drivable area will draw three-dimensional lanes on the interface, and the traffic lights and sign information will be displayed in a separate plane image. The object tracking will be reflected in the ID, the corresponding object ID information can be displayed on top of the 3D Box, and the motion prediction can be displayed as the future trajectory of the object. 

The planning information of the autonomous driving system is also reflected in the form of the future trajectory, and the control of the vehicle can be reflected by the length and Radian of the future trajectory, as well as the speed, direction, and, specific decision information of each moment. There is also an intuitive text display.

\begin{figure*}[htbp]
    \centering
    \subfigure[Perspective view]{
        \includegraphics[width=.31\linewidth]{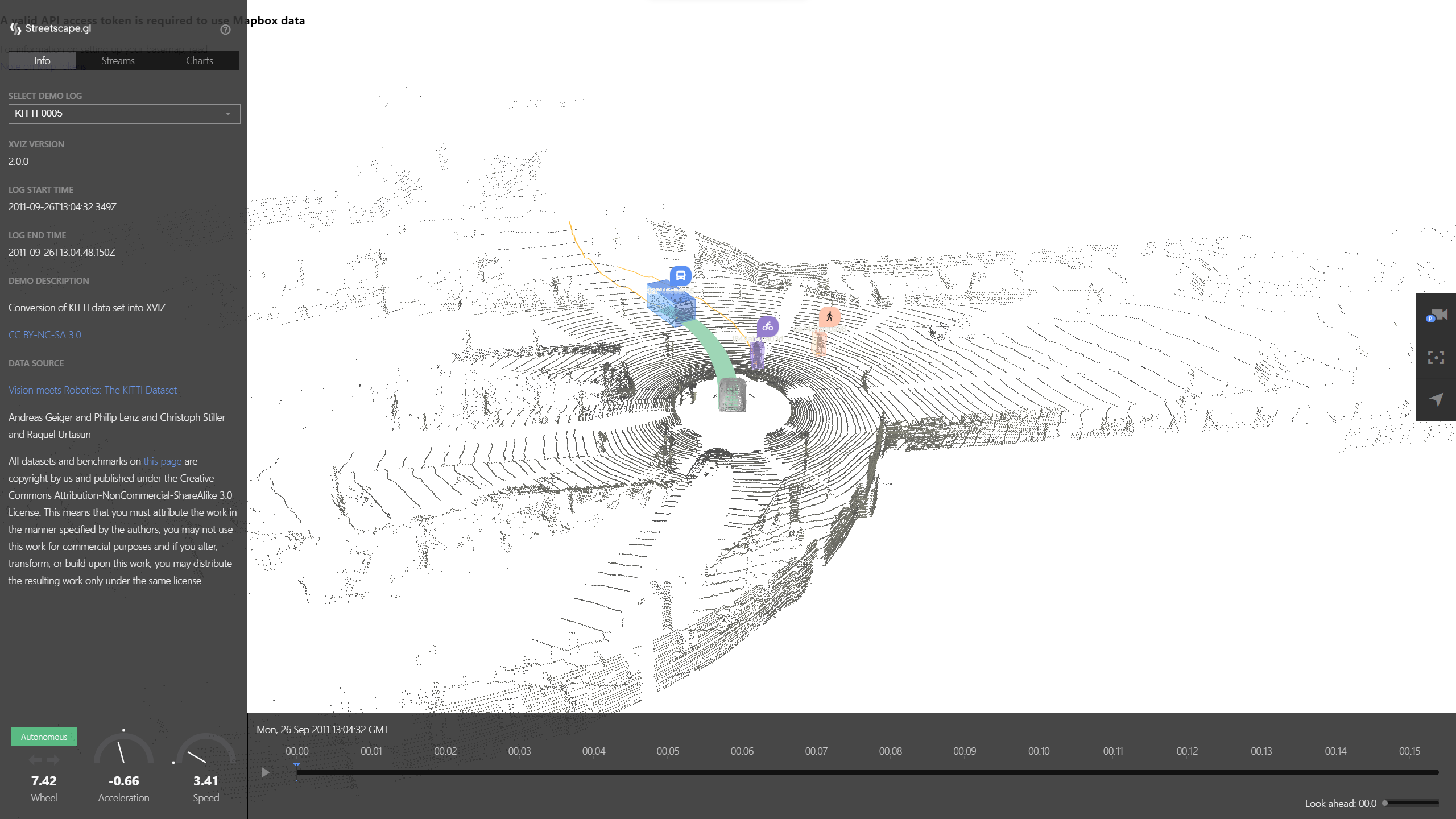}
        \label{}
    }
    \subfigure[Top-down view]{
        \includegraphics[width=.31\linewidth]{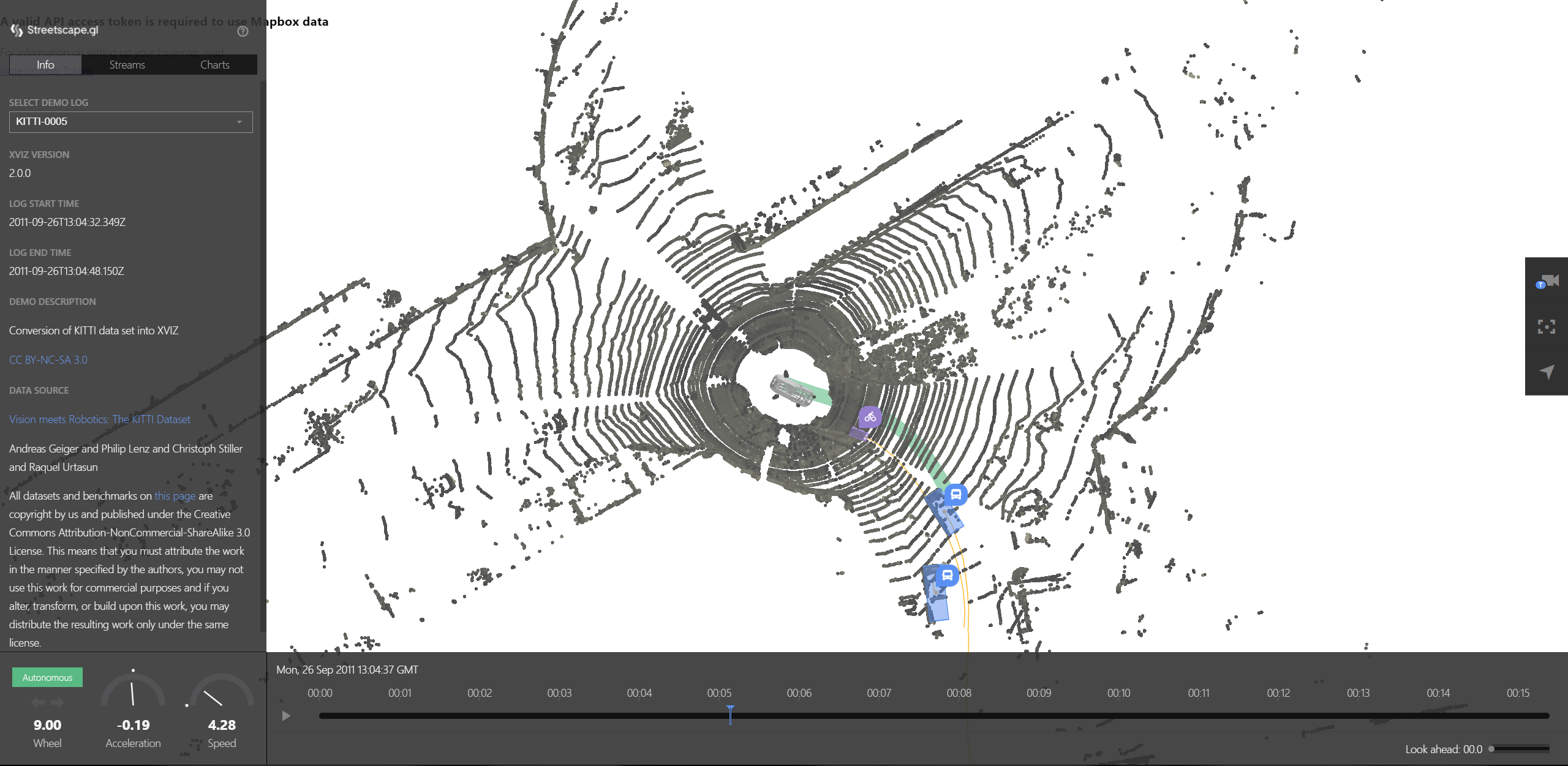}
        \label{webwiz}
    }
    \subfigure[Driver view]{
        \includegraphics[width=.31\linewidth]{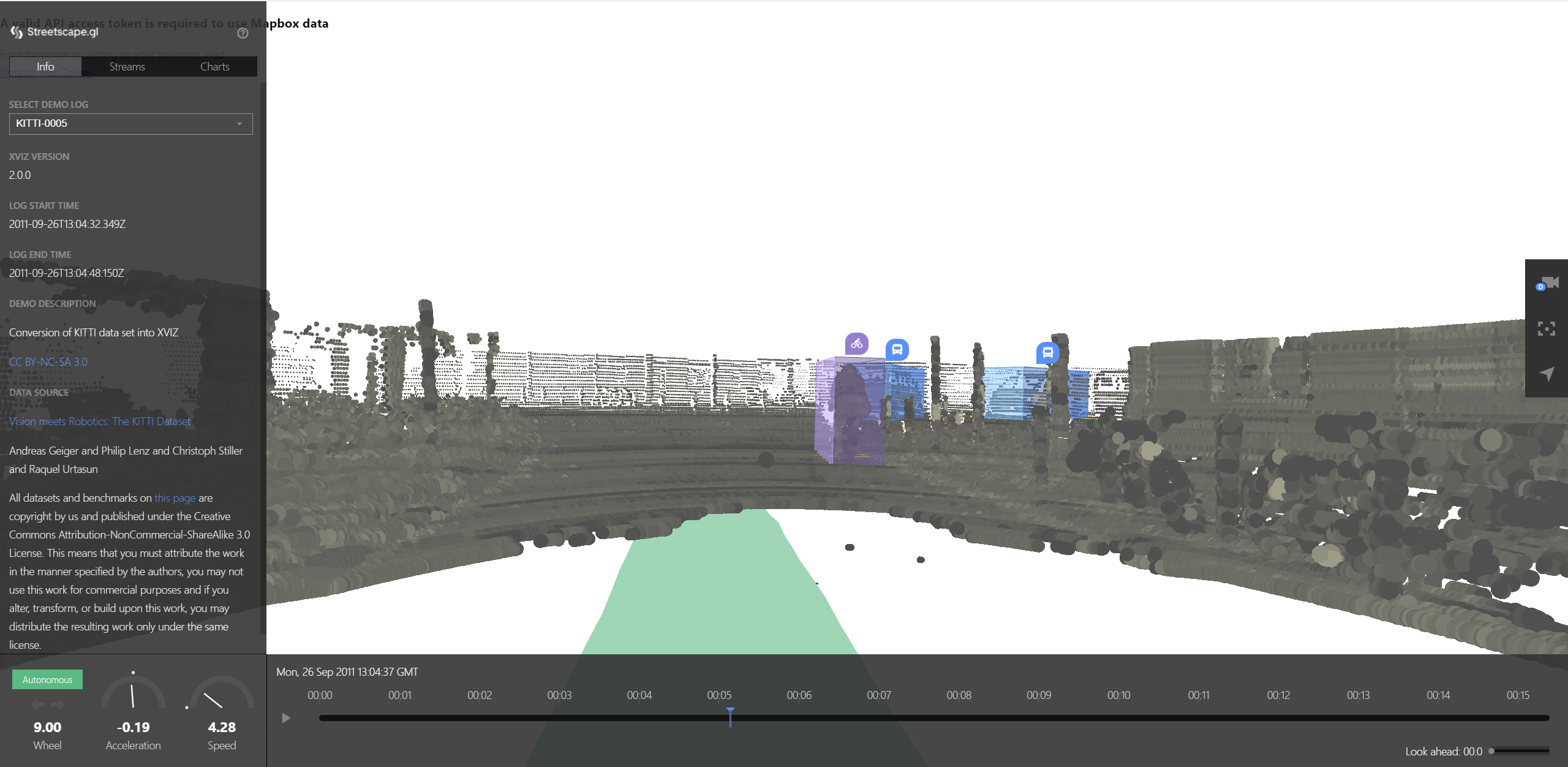}
        \label{webwiz}
    }
    \caption{Three default views of the visualization}
    \label{visualization}
\end{figure*}

In the visualization interface, we can filter the streams, move the camera, and can have three default views, perspective, top-down, and drive, as we showed in Figure.\ref{visualization}.

\section{Experimental setup}
\label{sec:setup}


We want to answer the following questions in the experiments:

\begin{itemize}
    \item Can the system simulates vehicle-road cooperative scenarios effectively?
    \item Is the system compatible with the KITTI data format?
    \item Is there a good visualization of the vehicle-road cooperative data?
\end{itemize}

To answer these questions, we carry out the functional tests and integration tests in this section. For the functional tests, we applied the KITTI dataset \cite{Geiger2013IJRR} to each layer of testing to test the compatibility of the system. The dataset structure is described in the Table.\ref{dataset structure}. For the integration test, we test the whole system and corresponding data pipeline as Figure.\ref{system} presents.

The system is deployed on a computer with Ubuntu 20.04.1, Carla version 0.9.13, CPU as Intel(R) Core(TM) i7-8700K CPU @ 3.70GHz, GPU as NVIDIA GeForce GTX 1080.

\begin{table}[ht]
    \centering
    \setcaptionwidth{.95\linewidth}
    \caption{Structure of 2011\_09\_26\_drive\_0005\_sync KITTI dataset}
    \begin{tabular}{|m{8em}|m{17em}|} \toprule
        \textbf{File/Folder Name} & \textbf{Description} \\
        tracklet\_labels.xml & Ground Truth value of the targets \\
        image\_0X & 4 sets of the same images with different color spaces \\ 
        oxts & The GPS/IMU information \\ 
        velodyne\_points & Point Cloud \\ \bottomrule
    \end{tabular}
    \label{dataset structure}
\end{table}

\subsection{Functional Test}

\begin{itemize}

\item{\textbf{{Simulation Layer}}}

For the Simulation Layer, we want to show its ability to simulate vehicle-road cooperate scenarios. Our simulation tools should be able to construct scenarios containing roads, the main vehicle, vehicles driving from opposite directions, and roadside devices, which is the basic vehicle-road cooperative scenario set. We should be able to set three types of road, straight road, curve road, and across. All the objects should be controlled synchronously by the simulation controller. As for the vehicle-road cooperate system, each main object, the main vehicle and the roadside devices should be able to output data separately.

\item{\textbf{{Sense and Perception Layer}}}
For the Sense and Perception Layer, we need to ensure the layer's compatibility with KITTI format data. The sensor module needs to output KITTI format data to provide simulation data of the vehicle-road cooperative system. We use the KITTI dataset described in the Table.\ref{dataset structure} to test the detection modules. We will input the KITTI point cloud data into the pre-trained algorithm model from OpenPCDet, and compare the output with the ground truth labels in each frame. As KITTI only provides the track-lets ground truth, we need to first extract the targets in each frame first. For the tracking  module, we input the ground truth labels into the tracking module. And here we expect the track-lets output to be the same format as the ground truth track-lets in the KITTI tracking  benchmark \cite{Geiger2012CVPR} as it's easier to process.

\item{\textbf{{Visualization Layer}}}
For the Visualization Layer, it needs to visualize the KITTI data. The point clouds and images should be displayed, as well as the ground truth targets and the GPS/IMU information. We may not care about the real position in the GPS, but we need to get the pose information of the vehicle from the IMU.

\end{itemize}

\subsection{Integration Test}
We do an integration test on the whole system to prove the effectiveness of the system, to show it can simulate and visualize the vehicle-road cooperative system. We will follow the data pipeline and finds the ability of the Visualization Layer on the vehicle-road cooperative data.

\section{Results}
\label{sec:results}


\subsection{Functional Test}

\begin{itemize}

\item{\textbf{{Simulation Layer}}}

We generate two scenarios shown in Figure.\ref{Scenes}. The main car is circled by blue. And the roadside device is circled by green. There's also a vehicle driving from another direction, which is covered by the architecture at first. And we can find three types of roads are generated successfully in the scenarios. These scenarios illustrate that the system has the ability to generate vehicle-road cooperative scenarios.

\begin{figure}[htbp]
    \centering
    \subfigure[Basic Scene 1]{
        \includegraphics[width=.5\linewidth]{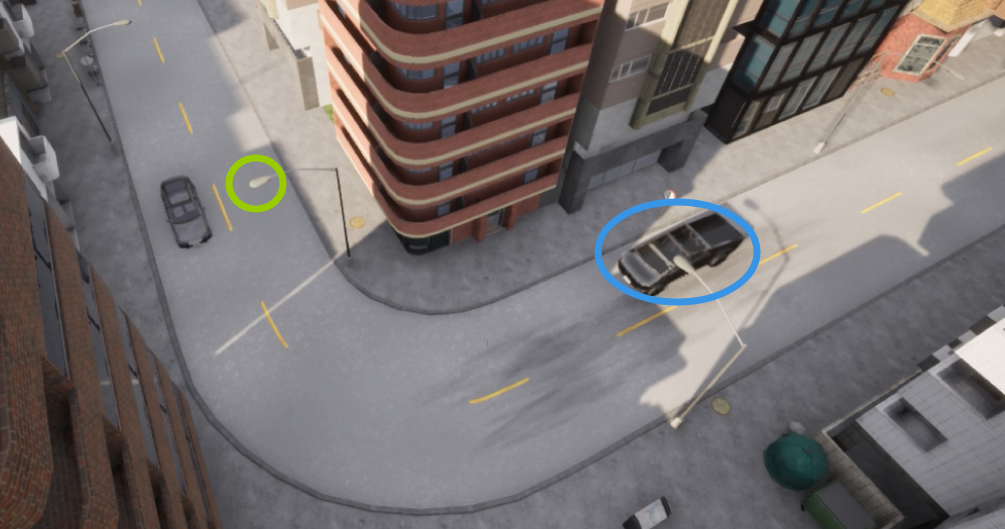}
        \label{scene1}
    }
    \subfigure[Basic Scene 2]{
        \includegraphics[width=.4\linewidth]{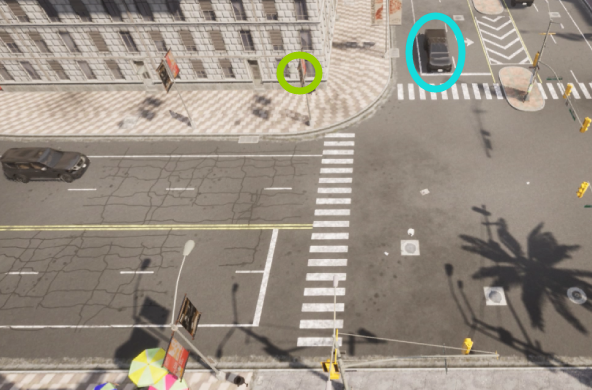}
        \label{scene2}
    }
    \setcaptionwidth{.9\linewidth}
    \caption{Basic Vehicle-road Cooperate Scene (The main vehicle is circled by blue, and the roadside device is circled by yellow)}\label{Scenes}
\end{figure}
\clearpage
\item{\textbf{{Sense and Perception Layer}}}

As we described in the system design, we mainly focus on point cloud detection. For the sensor module, we can respectively extract vehicle and roadside LiDAR point clouds, which are shown in the Figure.\ref{vehicleLiDAR}.

\begin{figure}[htb]
    \centering
    \subfigure[Vehicle LiDAR point cloud]{
        \includegraphics[width=.53\linewidth]{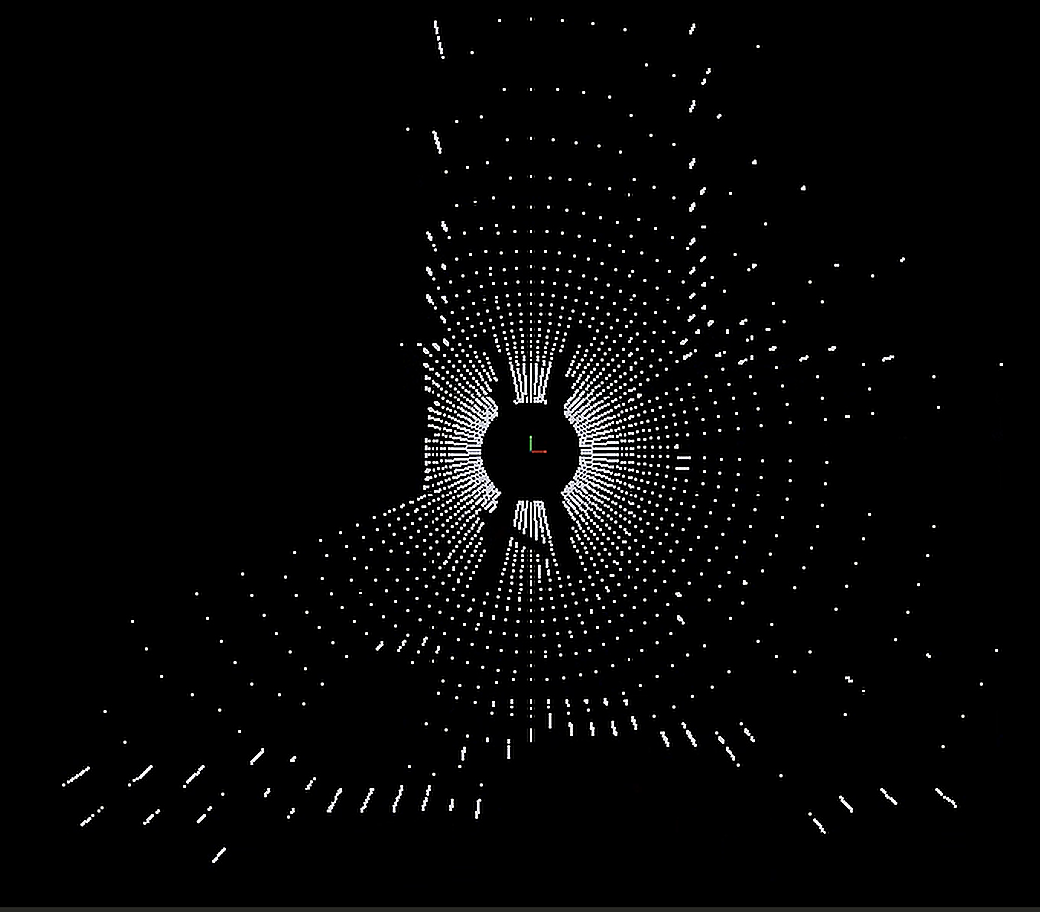}
        \label{vehicleLiDAR}
    }
    \subfigure[Road LiDAR point cloud]{
        \includegraphics[width=.4\linewidth]{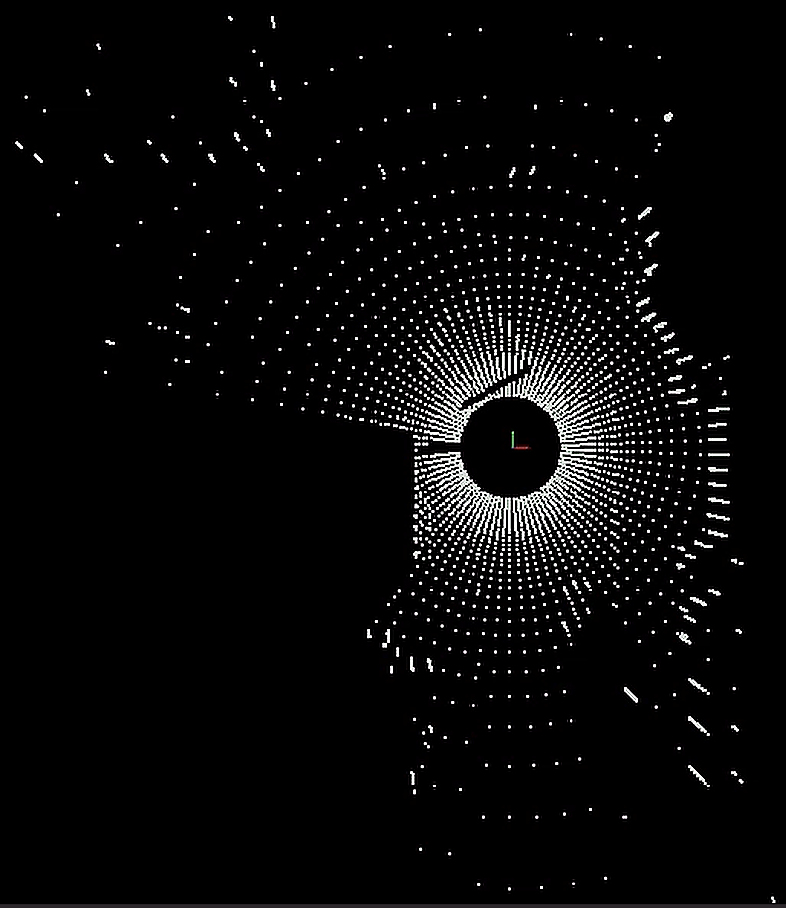}
        \label{roadLiDAR}
    }
    \caption{The LiDAR point clouds are extracted respectively from vehicle and roadside device}\label{LiDARs}
\end{figure}

For the detection module, we input the KITTI point clouds into the PointRCNN and Voxel R-CNN model. They detect most of the targets in the frames, only with some acceptable error on the targets' shapes and poses.

\begin{figure}[!ht]
    \centering
    \subfigure[The detection result of PointRCNN]{
        \includegraphics[width=.45\linewidth]{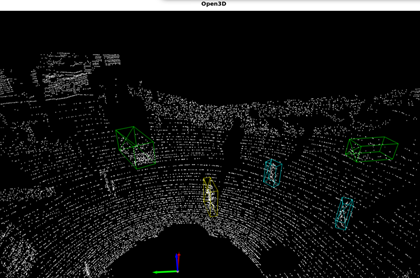}
        \label{pointrcnn}
    }
    \subfigure[Road LiDAR point cloud]{
        \includegraphics[width=.46\linewidth]{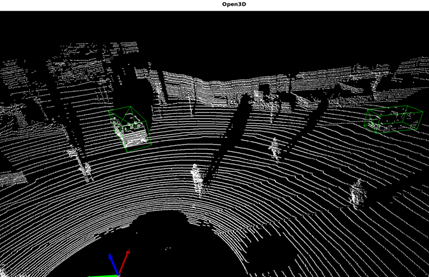}
        \label{voxelrcnn}
    }
    \caption{The detection result of Voxel R-CNN}\label{Detections}
\end{figure}

For the tracking  module, we respectively input the labels of each frame we extract from the ground truth track-lets and the detection results from the detection module. As AB3DMOT doesn't provide a good visualization tool for us to investigate the tracking result. We will discuss the result of this module in the Visualization Layer part.

\begin{figure}[!ht]
    \centering
    \includegraphics[width=.95\linewidth]{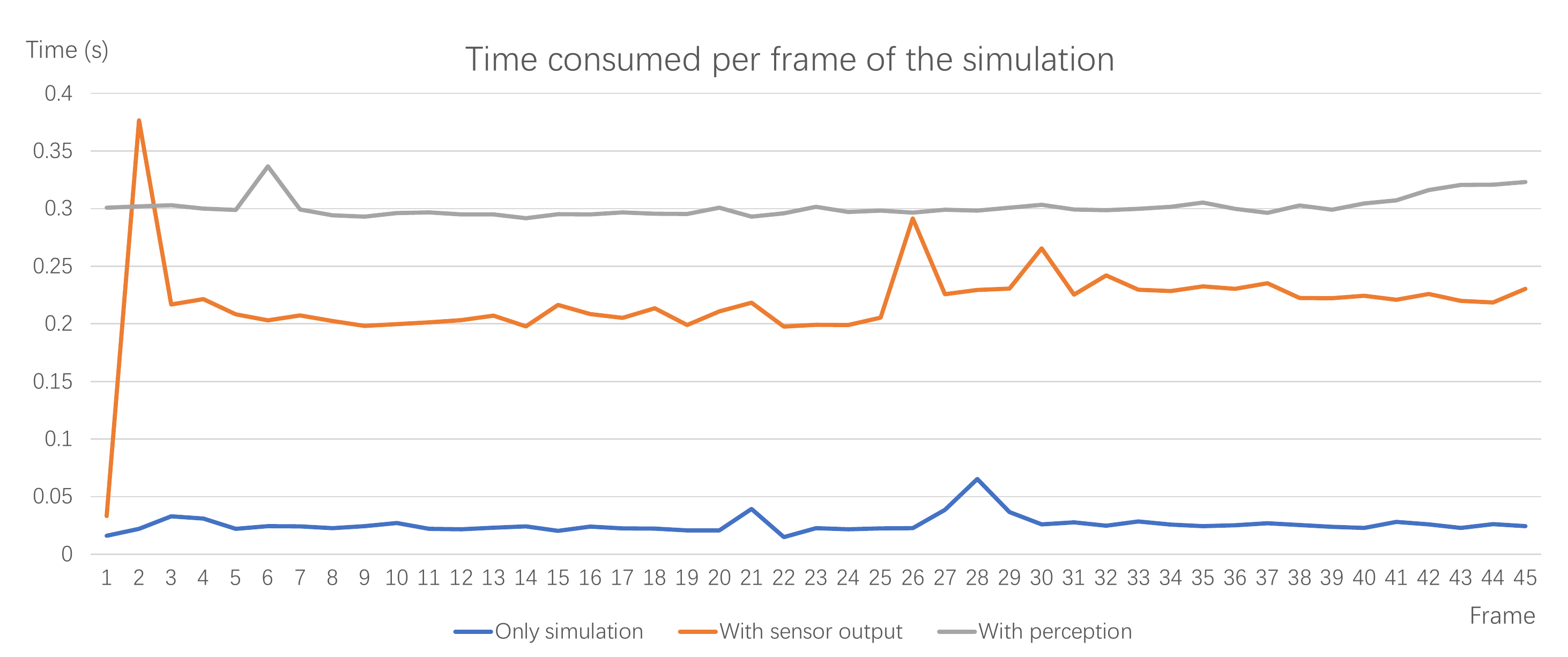}
    \caption{Time cost per frame of the simulation.}
    \label{time}
\end{figure}

As Figure.\ref{time} shows, that the time consumption is low for each frame of simulation under different configurations \cite{lan2022time}. Each frame of simulation cost average 0.025844052 seconds with only simulation. And average 0.21780155 seconds to save the disk. Average 0.309737757 seconds with real-time perception.

\item{\textbf{{Visualization Layer}}}

We first input the raw data of the KITTI dataset into our Visualization Layer. The result is shown in the figure.\ref{Visualization of raw KITTI dataset}. The visualization interface displays the data perfectly. All the information, point clouds, image streams, detected 3D boxes, and poses are displayed on the interface.

\begin{figure}[hp]
    \centering
    \includegraphics[width=.95\linewidth]{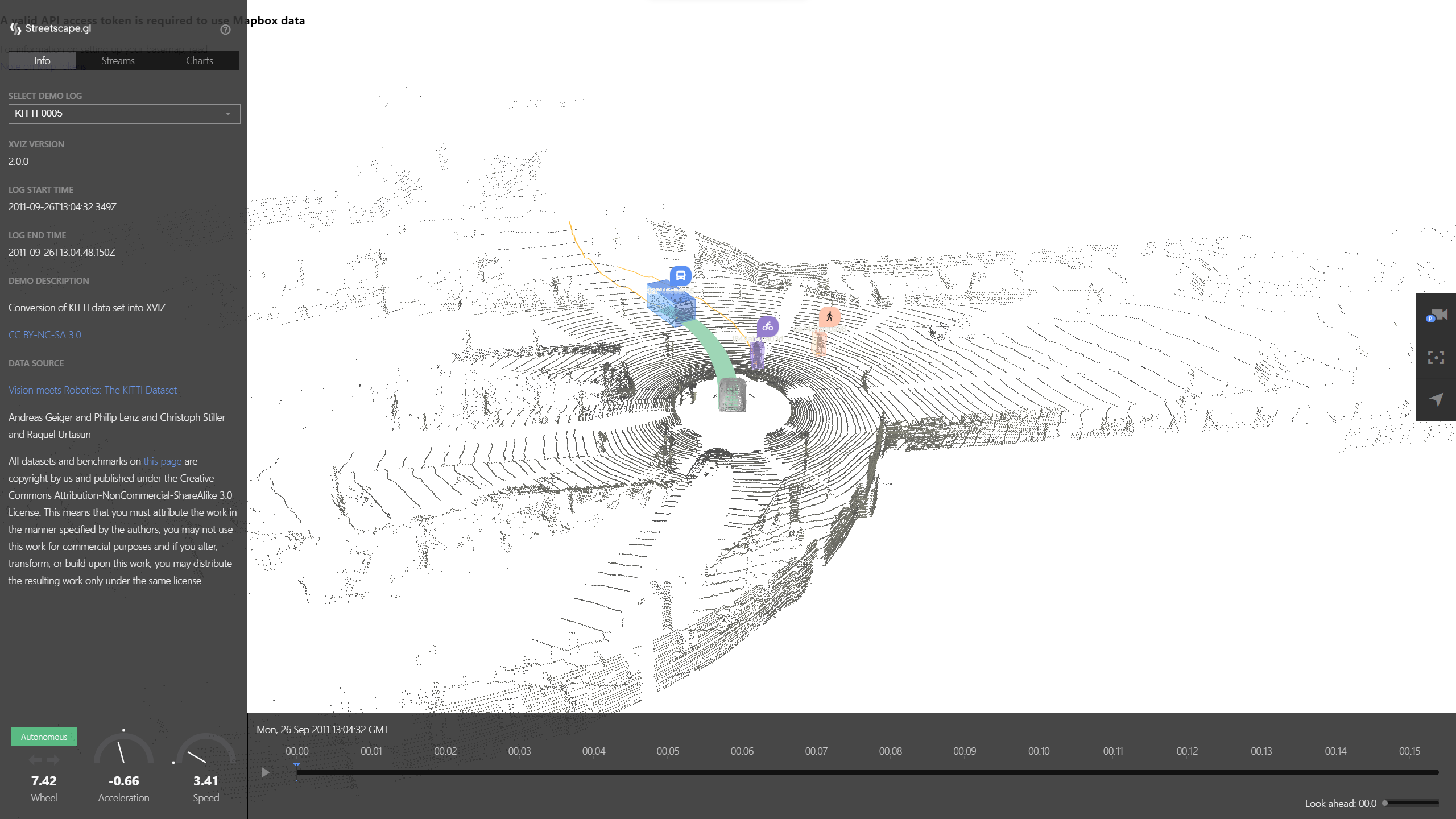}
    \caption{The 3D Visualization of the raw KITTI dataset}
    \label{Visualization of raw KITTI dataset}
\end{figure}

Then, we replace the ground truth track-lets with our tracking  results. For the detection result, most of the time, the 3D bounding boxes are just surrounding the bulge shape of the point cloud. And the tracking ids perform well in the consequent frames. We can say that our default modules are compatible with the KITTI data set and provide a high baseline for the perception layer.

\begin{figure}[htbp]
    \centering
    \subfigure[First frame]{
        \includegraphics[width=.8\linewidth]{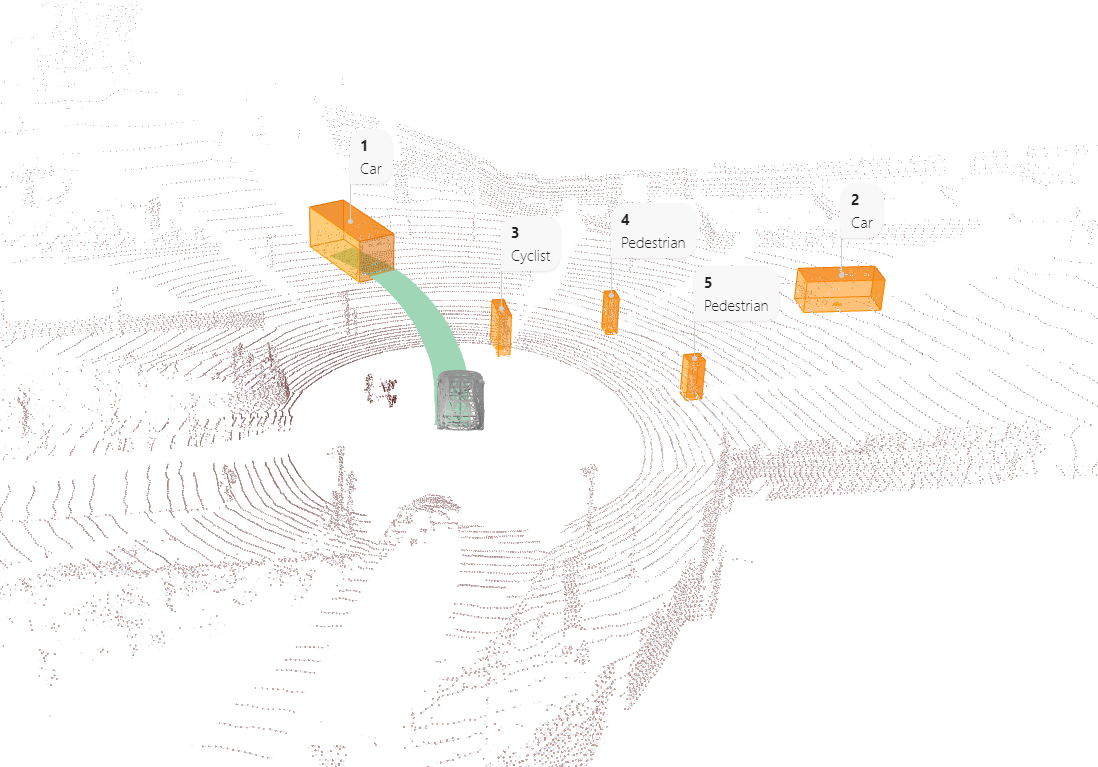}
        \label{frame0}
    }
    \subfigure[Second frame after a few frames of the first frame]{
        \includegraphics[width=.5\linewidth]{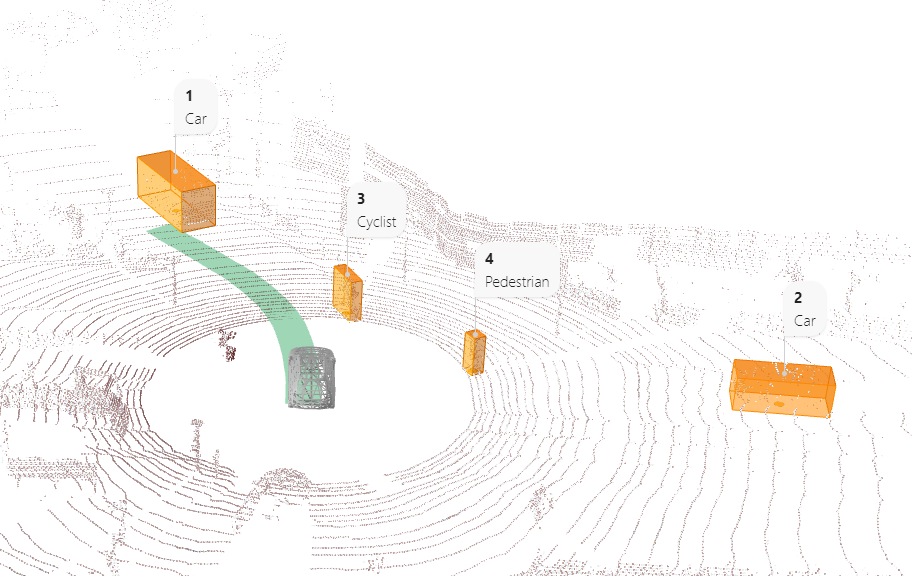}
        \label{frame1}
    }
    \subfigure[Third frame after a few frames of the second frame]{
        \includegraphics[width=.3\linewidth]{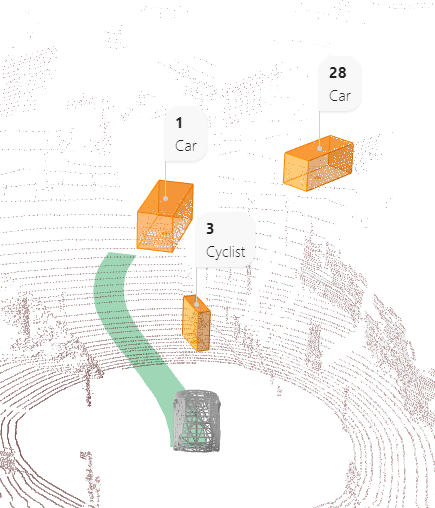}
        \label{frame2}
    }
    \caption{Tracking results among consequent frames}\label{}
\end{figure}
\end{itemize}

\begin{figure*}[htbp]
    \centering
    \subfigure[Detect result on vehicle LiDAR]{
        \includegraphics[width=.3\linewidth]{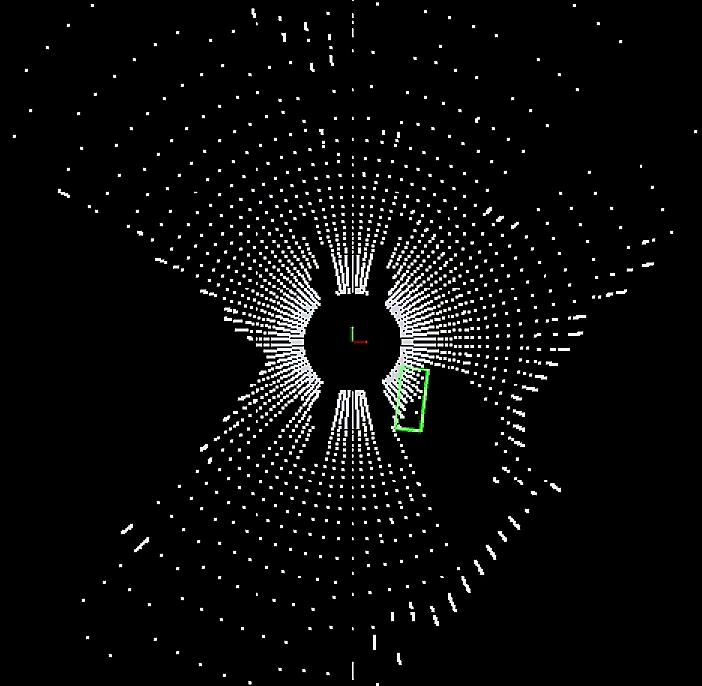}
        \label{vehicleLiDARRe}
    }
    \subfigure[Detect result on Road LiDAR]{
        \includegraphics[width=.3\linewidth]{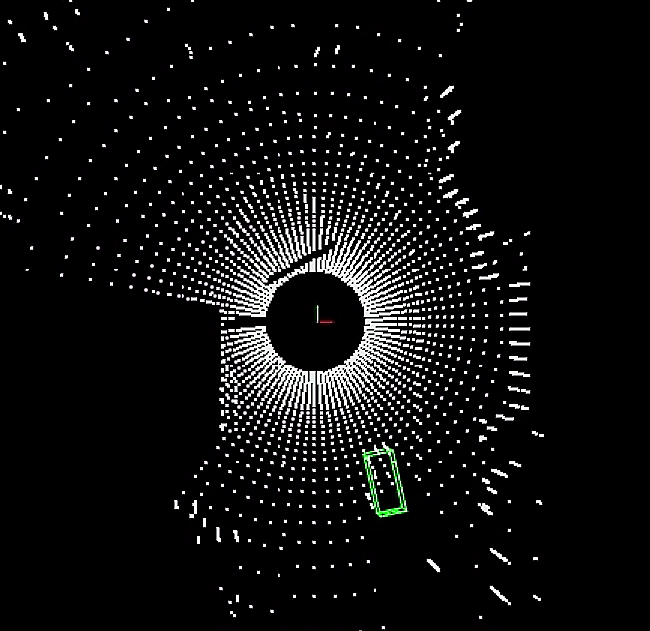}
        \label{roadLiDARRe}
    }
    \subfigure[Decision-level merge result of the cooperative system]{
        \includegraphics[width=.305\linewidth]{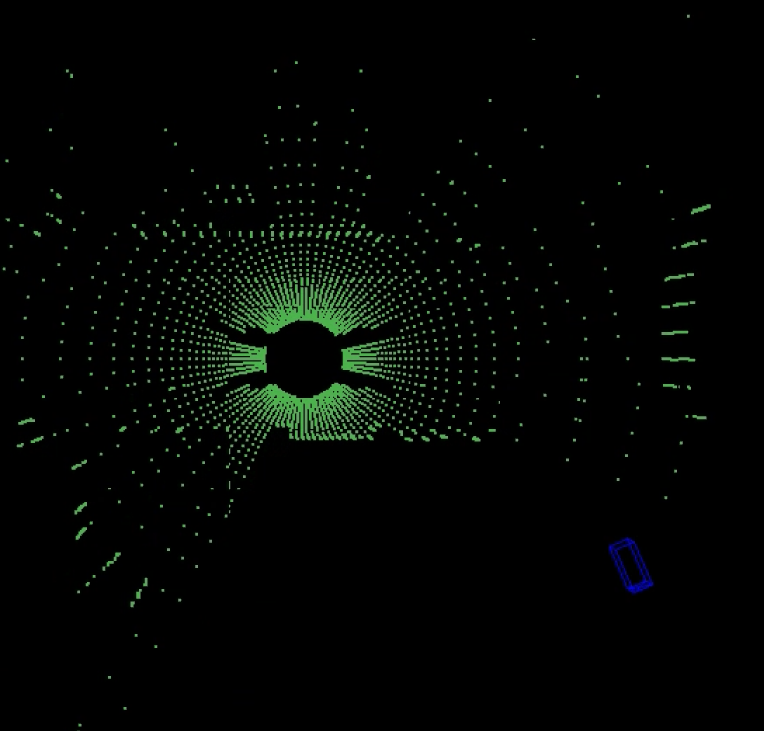}
        \label{merge}
    }
    \caption{The detection results on LiDAR point clouds}\label{LiDARRe}
\end{figure*}

\subsection{Integration Test}
In the integration test, we follow the system's data pipeline. We first extract KITTI format data from Simulation Layer. Then, get the results from the Sense and Perception Layer, and process them in the Application Layer. Finally, put all the data into Visualization Layer. As Figure.\ref{LiDARRe} shows. We put the sensors data into the detection module, the detection results are OK for the vehicle LiDAR data. But for the roadside LiDAR, the continuity of target detection is poor. We consider that maybe the pre-trained model is only trained on real-world LiDAR data and is not compatible with virtual LiDAR data. Or the LiDAR settings are different, such as different LiDAR scan angles, scan line density, placement height, etc. Or the model is trained on vehicle LiDAR data, and simply can't detect from roadside view. Researchers can try to replace the algorithm model or retrain the model with virtual radar data to improve the results.

\begin{figure}[!ht]
    \centering
    \includegraphics[width=0.95\linewidth]{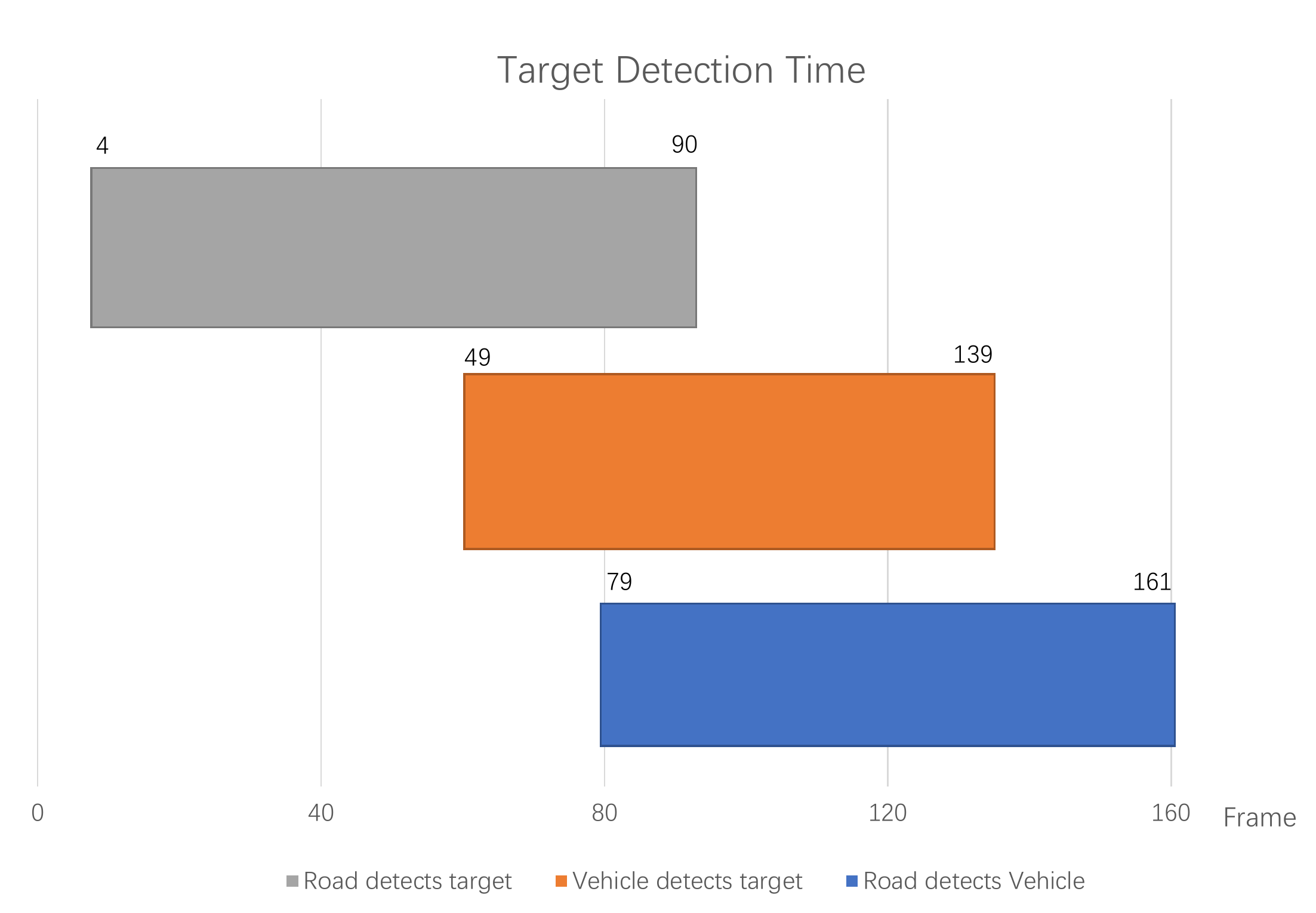}
    \caption{The detection result from the vehicle and roadside among frames.}
    \label{detect}
\end{figure}

Though the detection score may not be very well, we can still investigate the power of vehicle-road cooperation. As Figure.\ref{detect} shows, that the roadside device detects the targets at first. The target goes into the vehicle's detection range, and then the vehicle comes into the roadside device's detection range. The vehicle covers the target which makes the roadside device can't track the target. And finally, the target leaves the detection range of the vehicle, and the vehicle leaves the roadside device's detection range. We can find that vehicle-road cooperation can help the vehicle detects the targets in advance.

The tracking module performs well on the detection results. Then we merge the information in the Application Layer and put all the data into the Visualization Layer. All the information is displayed on the interface as expected. For the vehicle-r.oad cooperative data, we can see that the vehicle data and roadside data are displayed in different colors, which separate them. Besides, we can control the visibility of each data stream in the left menu. So that we can investigate the differences in the vehicle-road cooperative system.

\begin{figure}[htbp]
    \centering
    \includegraphics[width=0.95\linewidth]{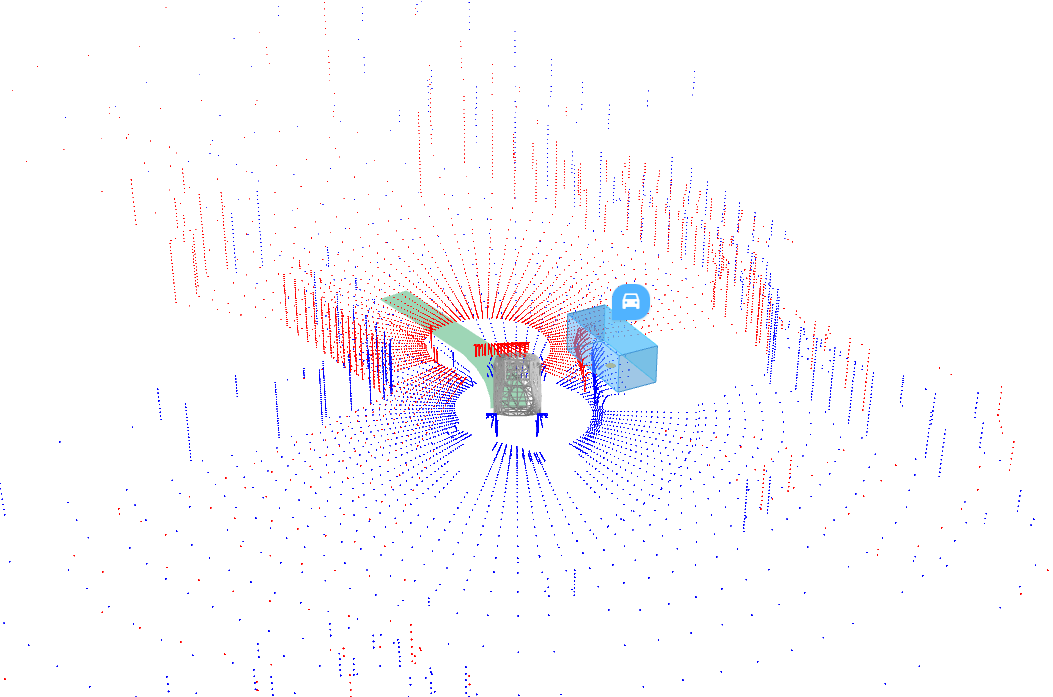}
    \caption{The 3D Visualization of the vehicle-road cooperative simulation}
    \label{Visualization of Vehicle-road cooperative}
\end{figure}

The result shows that our system can simulate vehicle-road cooperative scenarios and extract vehicle-road cooperative data. The system is also compatible with the KITTI data format, which enables it to support many algorithms that are tested on KITTI data. The system also provides a good visualization of the vehicle-road cooperative data so that the researcher can easily interact with the interface and investigate the differences between the vehicle-road cooperative system.

\section{Discussion}
\label{sec:discussion}


In this paper, we present a Vehicle-road Cooperative Simulation and Visualization system, a generalized framework and tool for research and development of vehicle-road cooperative technologies. It contains Simulation Layer, Sense Layer, Application Layer, and Visualization Layer, providing default modules in each layer that are changeable. And we construct the data pipeline between layers to enable researchers to test and evaluate the vehicle-road cooperative autonomous driving systems. The system is compatible with the KITTI data format and provides vehicle-road cooperative data with the format of an extension from KITTI. We envision that our system will provide a framework to do research on the vehicle-road cooperative system. And we will provide more vehicle-road cooperative scenarios, develop more modules in each layer, and develop more evaluation tools in the future \cite{lan2022semantic}.










\bibliographystyle{IEEEtran}
\bibliography{bibliography}

\end{document}